# Optimal Surface Segmentation with Convex Priors in Irregularly Sampled Space

Abhay Shah[a], Michael D. Abramoff[a,b], Xiaodong Wu[a,c]

[a]Department of Electrical and Computer Engineering, The University of Iowa, Iowa City, IA, 52242, USA
[b]Department of Ophthalmology and Visual Sciences University of Iowa, Iowa City, IA, 52242, USA
[c]Department of Radiation Oncology, University of Iowa, Iowa City, IA 52242, USA

## Abstract

Optimal surface segmentation is a state-of-the-art method used for segmentation of multiple globally optimal surfaces in volumetric datasets. The method is widely used in numerous medical image segmentation applications. However, nodes in the graph based optimal surface segmentation method typically encode uniformly distributed orthogonal voxels of the volume. Thus the segmentation cannot attain an accuracy greater than a single unit voxel, i.e. the distance between two adjoining nodes in graph space. Segmentation accuracy higher than a unit voxel is achievable by exploiting partial volume information in the voxels which shall result in non-equidistant spacing between adjoining graph nodes. This paper reports a generalized graph based multiple surface segmentation method with convex priors which can optimally segment the target surfaces in an irregularly sampled space. The proposed method allows non-equidistant spacing between the adjoining graph nodes to achieve subvoxel segmentation accuracy by utilizing the partial volume information in the voxels. The partial volume information in the voxels is exploited by computing a displacement field from the original volume data to identify the subvoxel-accurate centers within each voxel resulting in non-equidistant spacing between the adjoining graph nodes. The smoothness of each surface modeled as a convex constraint governs the connectivity and regularity of the surface. We employ an edge-based graph representation to incorporate the necessary constraints and the globally optimal solution is obtained by computing a minimum $s$-$t$ cut. The proposed method was validated on 10 intravascular multi-frame ultrasound image datasets for subvoxel segmentation accuracy. In all cases, the approach yielded highly accurate results. Our approach can be readily extended to higher-dimensional segmentations.

---

## 1. Introduction

Optimal surface segmentation method for 3-D surfaces representing object boundaries is widely used in image understanding, object recognition and quantitative analysis of volumetric medical images (Li et al., 2006; Abra`moff et al., 2010; Withey and Koles, 2008). The optimal surface segmentation technique (Li et al., 2006) has been extensively employed for segmentation of complex objects and surfaces, such as knee bone and cartilage (Yin et al., 2010; Kashyap et al., 2013), heart (Wu et al., 2011; Zhang et al., 2013), airways and vessels tress (Liu et al., 2013; Bauer et al., 2014), lungs (Sun et al., 2013), liver (Zhang et al., 2010), prostate and bladder (Song et al., 2010), retinal surfaces (Garvin et al., 2009; Lee et al., 2010) and fat water decomposition (Cui et al., 2015). The segmentation problem is transformed into an energy minimization problem (Li et al., 2006; Ishikawa, 2003; Boykov et al., 2001). A graph is then constructed, wherein the graph nodes correspond to the center of evenly distributed voxels (equidistant spacing between adjoining nodes). Edges are added between the nodes in the graph to correctly encode the different terms in the energy function. The energy function can then be minimized using a minimum *s-t* cut (Li et al., 2006; Boykov and Kolmogorov, 2004). The resultant minimum *s-t* cut corresponds to the surface position of the target surface in the voxel grid.

The method requires appropriate encoding of primarily the following three types of energy terms (Song et al., 2013; Shah et al., 2015) into the graph construction. The data term (also commonly known as the data cost term) which measures the inverse likelihood of all voxels on a surface, a surface smoothness term (surface smoothness constraint) which specifies the regularity of the target surfaces and a surface separation term (surface separation constraint) which governs the feasible distance between two adjacent surfaces. A detailed description of the energy terms is provided in Section 2.1. Various types of surface smoothness and surface separation constraints are used for simultaneous segmentation of multiple surfaces. Optimal surface detection method (Li et al., 2006; Wu and Chen, 2002) uses hard smoothness constraints that are a constant in each direction to specify the maximum allowed change in surface position of any two adjacent voxels on a feasible surface. It uses hard surface separation constraints to specify the minimum and maximum allowed distances between a pair of surfaces. Methods employing trained hard constraints (Garvin et al., 2009), use prior term to penalize local changes in surface smoothness and surface separation. The constraints can also be modeled as a convex function (convex smoothness constraints) as reported in Ref. (Song et al., 2013; Dufour et al., 2013). Furthermore, a truncated convex function (truncated convex constraints) may also be used to model the surface smoothness and surface separation constraints (Kumar et al., 2011; Shah et al., 2014, 2015) to segment more complex surfaces but does not guarantee global optimality. A truncated convex constraint enforces a convex function based penalty with a bound on the maximum possible penalty.

However, since volumetric data is typically represented as an orthogonal matrix of intensities, the surface segmentation cannot achieve a precision greater than a single unit voxel, i.e. the distance between two adjoining nodes in the graph space. Accuracy higher than a single unit voxel (subvoxel accuracy) can be attained by exploiting partial volume effects in the image volumes (Abr`amoff et al., 2014; Malmberg et al., 2011) which leads to non-equidistant spacing between the adjoining graph nodes resulting in an irregularly sampled space. Volumetric images are obtained by discretizing the continuous intensity function uniformly sampled by sensors, resulting in partial volume effects (Shannon, 1949; Trujillo-Pino et al., 2013). Partial volume effects are inherent in images as voxels 'combine' partial information from various features (such as tissues) of the imaged object. The spatial resolution in images is limited by the detector/sensor design and by the reconstruction process, which results in 3-D image blurring introduced by the finite spatial resolution of the imaging system (Soret et al., 2007). Mathematically, the finite resolution effect is described by a 3-D convolution operation, where the image is formed by the convolution of the actual source with the 3-D point spread function of the imaging system, which causes spillover between regions. The signal intensity in each voxel is the mean of signal intensities of the underlying tissues included in that voxel. The ignored partial volume information can be utilized by computing a displacement field directly from the volumetric data (Abra`moff et al., 2014) to identify the subvoxel-accurate location of the centers within each voxel, thus requiring a generalized construction of the graph with non-equidistant spacing between orthogonal adjoining nodes (irregularly sampled space). Increased subvoxel segmentation accuracy attained by exploiting the partial volume effects has the potential for better diagnosis and treatment of disease.

The optimal surface segmentation technique employing the different types of smoothness constraints as discussed above is not capable of efficiently segmenting surfaces with subvoxel accuracy in a volume which requires segmentation in a grid comprising of non-uniformly sampled voxels where the spacing between the orthogonally adjoining nodes is not equidistant.

To address this problem, the subvoxel accurate graph search method (Abra`moff et al., 2014) was developed to simultaneously segment multiple surfaces in a volumetric image by exploiting the additional partial volume information in the voxels. A displacement field is computed from the original volumetric data. The method first creates the graph using the conventional optimal surface segmentation method (Li et al., 2006), then deforms it using a displacement field and finally adjusts the inter-column edges and inter-surface edges to incorporate the modification of these constraints. Specifically, such a deformation shall result in non-equidistant spacing between the adjoining nodes which may be considered equivalent to a generalized case of a cube volume formed by non-uniform sampling along the $z$ dimension for the purposes of 3-D surface segmentation. The method demonstrated achievement of subvoxel accuracy compared to the traditionally used optimal surface segmentation method (Li et al., 2006). An example is shown in Fig. 1. However, the method employs hard surface smoothness which does not allow flexibility in constraining surfaces. Specifically, the previous approach was not capable of incorporating a convex surface smoothness constraint in the graph with non-equidistant spacing between adjoining nodes.

Our main contribution is extension of the framework presented in Ref.(Abr`amoff et al., 2014) to incorporate convex surface smoothness/separation constraints for multiple surface segmentation in irregularly sampled space. The proposed method is a generalization of the graph based optimal surface segmentation with convex priors (Song et al., 2013) in the regularly sampled space. Consequently, the graph constructed in the regularly sampled space forms a special case in the irregularly sampled space framework where the spacing between the adjoining nodes is set to be a constant (equidistant). The use of convex priors allows for incorporation of many different prior information in the graph framework as discussed previously while attaining subvoxel accuracy. Unlike the subvoxel accurate graph search method (Abra`moff et al., 2014), the proposed method does not require a two step process to create the graph by the conventional method and then readjust the edges, but instead provides a one step function to add edges between nodes from two neighboring columns to incorporate the convex prior.

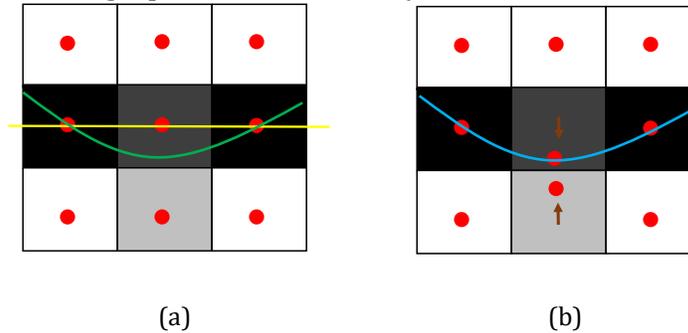

(a)                                    (b)

Figure 1: Example of a 3×3 voxel grid to demonstrate subvoxel accuracy. Each voxel is represented by a red node in the graph space. (a)Graph nodes with equidistant spacing between them. True subvoxel accurate surface is shown in green. The segmented surface using optimal surface segmentation method with hard constraints is shown in yellow. (b) The displacement field derived from the grid is applied to the central nodes displacing the centers to exploit the information from the partial volume effect shown by brown arrows. The resultant segmentation with the subvoxel accurate graph search is shown in blue.

Subvoxel surface segmentation methods employing adaptive grids (Lang et al., 2014) and located cuts (Malmberg et al., 2011) have also been used to segment surfaces with subvoxel precision. The adaptive grid methodology (Lang et al., 2014) requires a pre-segmentation of the target surfaces and generates an application specific grid, wherein, the graph nodes are only placed in the region of interest between the inner and outer surfaces by performing flattening of the surfaces using a regression model. The surfaces are then segmented using the optimal surface segmentation method (Li et al., 2006). The sub pixel segmentation method as described in Ref. (Malmberg et al., 2011), utilizes an initial segmentation to create fuzzy vertices in the graph using a distance transform. Utilizing the information from the fuzzy vertex segmentation, a located cut for the boundary of the vertex segmentation is then derived to compute the final segmentation. Both methods essentially make local adjustments and improvements to the segmentation in the regularly sampled space, while the proposed method computes the globally optimal solution from the graph constructed in the irregularly sampled space.

In addition, the adaptive moving grid has been used for solving partial differential equations (PDEs) (Budd et al., 2009). The grid adaptivity also finds its application in the quadtree and octree methods for improving resolution locally in a hierarchical data representation (Samet, 1988).

Note that a straightforward way to solve the problem is to simply upsample the columns and directly apply the graph search method, which increases the graph size proportional to the factor of upsampling, thus resulting in very high computation time and is dependent on determination of the minimum scale of subvoxel-accurate segmentation. The proposed method does not require any such upsampling and is capable of segmenting the target surfaces in the available resolution with subvoxel accuracy. Additionally, the proposed method does not introduce additional parameters in the formulation in comparison with graph search method (Li et al., 2006).

In the following sections, we briefly explain the formulation for the optimal surface segmentation method in the regularly sampled space, explain the formulation and description of our novel graph construction to incorporate the convex smoothness constraints in the irregularly sampled space. Next, the evaluation is performed on intravascular multi-frame ultrasound image datasets for validation and applicability of the method to demonstrate subvoxel segmentation accuracy compared to optimal surface segmentation method with convex priors in regularly sampled space (Song et al., 2013). Finally, the proof for correctness of graph construction to model the convex surface smoothness constraints is presented in Appendix A and B.

## 2. Methods

### 2.1. Problem Formulation and Energy Function

The problem formulation for the widely used optimal surface segmentation methods (Li et al., 2006; Wu and Chen, 2002; Song et al., 2013) is described as follows. Consider a volume $I(x,y,z)$ of size $X \times Y \times Z$. A surface is defined as a function $S(x,y)$, where $x \in \mathbf{x} = \{0,1,...X-1\}$, $y \in \mathbf{y} = \{0,1,...Y-1\}$ and $S(x,y) \in \mathbf{z} = \{0,1,...Z-1\}$. It is worth noting that the center of voxels are uniformly sampled. Each $(x,y)$-pair corresponds to a voxel *column* $\{(I(x,y,z)|z = 0,1,...,Z-1\}$. We use $a$ and $b$ to denote two columns corresponding to two neighboring $(x,y)$-pairs in the domain $\mathbf{x} \times \mathbf{y}$ and $N_s$ to denote the set of pairs of neighboring columns. The function $S(a)$ can be viewed as labeling for $a$ with the label set $\mathbf{z}$ ($S(a) \in \mathbf{z}$). For simultaneously segmenting $\lambda (\lambda \geq 2)$ distinct but interrelated surfaces, the goal of the problem is to seek the globally optimal surfaces $S_i(a)$, where $i = 1,2,\ldots\lambda$ in $I$ with minimum separation $d_{j,j+1}$ where $j = 1,2,\ldots\lambda - 1$ between each adjacent pair of surfaces $S_j$ and $S_{j+1}$.

The problem is transformed into an energy minimization problem. The energy function $E(S)$ takes the following form as shown in Eqn. (1):

$$E(S) = \sum_{i=1}^{\lambda} \sum_{a \in \mathbf{x} \times \mathbf{y}} ( \sum_{(a,b) \in N_s} D_i(S_i(a)) + \sum V_{ab}(S_i(a), S_i(b)))$$
$$+ \sum_{i=1}^{\lambda-1} \sum_{a \in \mathbf{x} \times \mathbf{y}} H_a(S_{i+1}(a), S_i(a)) \tag{1}$$

The data cost term $\mathrm{P}_{a \in \mathbf{x} \times \mathbf{y}} D_i(S_i(a))$ measures the total cost of all voxels on a surface $S_i$, where $D_i$ measures the inverse probability of a voxel belonging to surface $S_i$. The surface smoothness term $\mathrm{P}_{(a,b) \in N_s} V_{ab}(S_i(a), S_i(b))$ constrains the connectivity of a surface in 3-D and regularizes the surface. Intuitively, this defines how rigid the surface is. The surface separation term $H_a(S_i(a), S_{i+1}(a))$ constrains the distance of surface $S_i$ to $S_{i+1}$. The energy function is appropriately encoded in a graph. A minimum $s$-$t$ cut is then computed on the graph to get solutions for the target surfaces $S_i$'s.

Typically graph construction is done with equidistant spacing between the adjoining nodes (regularly sampled space). Our main contribution is to allow for optimal surface segmentation in the irregularly sampled space with convex surface smoothness/separation constraints by allowing non-equidistant spacing between the nodes.

We formulate the multiple surface segmentation problem in a similar manner for the irregularly sampled space. Consider a volume $\Gamma(x,y,\tilde{z})$ where $x \in \mathbf{x} = \{0,1,\ldots X - 1\}$, $y \in \mathbf{y} = \{0,1,\ldots Y - 1\}$ and $\tilde{z} \in \mathsf{R}$. Each $(x,y)$-pair corresponds to a *column* $\{(\Gamma(x,y,\tilde{z}) | \tilde{z} \in \mathsf{R}$, denoted by $col(x,y)$. Assume each $col(x,y)$ has exactly $\tilde{Z}$ elements obtained by sampling strictly in the increasing order along the $\tilde{z}$ direction which are indexed by $\{0,1,\ldots\tilde{Z}-1\}$ along $col(x,y)$. This yields a volumetric image $I(x,y,z)$ of size $X \times Y \times \tilde{Z}$, where $x \in \mathbf{x} = \{0,1,\ldots X - 1\}$, $y \in \mathbf{y} = \{0,1,...Y-1\}$ and $\tilde{z} \in \tilde{\mathbf{z}} = \{0,1,...\tilde{Z}-1\}$, which allows for non-equidistant spacing between two adjacent elements in the column. As discussed previously $a$ and $b$ are used to denote two neighboring columns. For ease of understanding, we assume $\tilde{Z} = Z$ for the remainder of this paper.

Note, for purposes of the experiments in this paper, relaxation of equidistance constraint concerns the z axis only. As the image domain we consider is an x-y grid, we thus only relax the equidistance constraint along the z-axis. It is possible and would be useful to relax the equidistance constraint in the x- and y-axes if the image domain is defined on a meshed simple surface, that is, the sought surface is monotone to the meshed surface. However, to avoid the interference, we may restrict to move the center point around within each voxel.

We define a mapping function for each column $a$ as $L_a : \{0,1,...Z-1\} \rightarrow \mathbb{R}$ which maps the index of sampled points in $I(a,z)$ to $\Gamma(a,\tilde{z})$. For example, $L_a(i)$ denotes the $\tilde{z}$ coordinate of the $i+1$-th sample along column $a$, and $L_a(i+1) > L_a(i)$ because of the strictly increasing order of sampling along column $a$. An example is shown in Fig. 2. Further, a surface labeling for column $a$ is defined as $S(a)$, where $S(a) \in \mathbf{z} = \{0,1,...Z-1\}$. The function $L_a(S(a))$ defines the "physical" location (the $\tilde{z}$ coordinate) of surface $S$ at column $a$. For simultaneously segmenting $\lambda$ ($\lambda \geq 2$) surfaces, the goal of the problem is to seek the surface labeling $S_i(a)$ on all columns in $I$ for each surface $S_i$, where $i = 1,2...\lambda$, with minimum separation $d_{j,j+1}$ where $j = 1,2,...\lambda-1$ between adjacent pair of surfaces. It is to be noted, that the surfaces are ordered, i.e, $L_a(S_{i+1}(a)) \geq L_a(S_i(a))$. The corresponding energy function for this formulation is shown in Equation 2:

$$
\begin{aligned}
E(S) = &\sum_{i=1}^{\lambda} \Big( \sum_{a \in \mathbf{x} \times \mathbf{y}} D_i(L_a(S_i(a))) \\
&+ \sum_{(a,b) \in N_i} V_{ab}(L_a(S_i(a)), L_b(S_i(b))) \\
&+ \sum_{i=1}^{\lambda-1} \sum_{a \in \mathbf{x} \times \mathbf{y}} H_a(L_a(S_{i+1}(a)), L_a(S_i(a)))
\end{aligned}
\qquad (2)
$$

Herein, the surface smoothness term is modeled as a convex function as shown in Equation (3).

$$V_{ab}(L_a(S_i(a)), L_b(S_i(b))) = \psi(L_a(S_i(a)) - L_b(S_i(b)))$$ (3)

where, $\psi(.)$ is a convex function, and without loss of generality, we assume that $\psi(0) = 0$ Wu and Chen (2002).

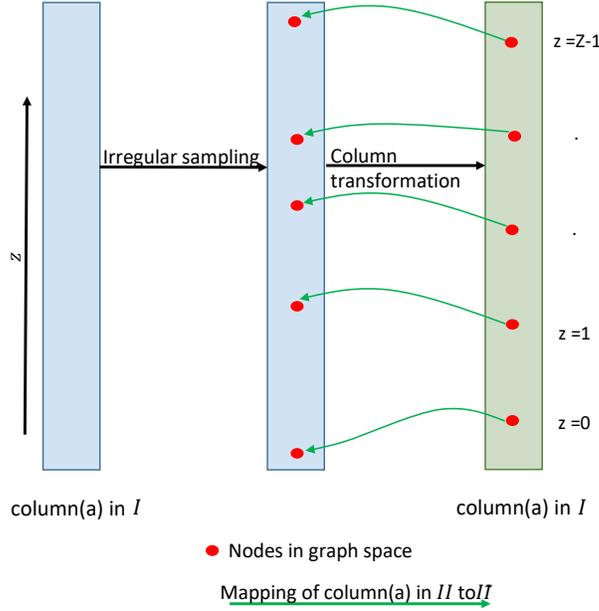

Figure 2: Example of column structure for irregularly sampled space using mapping function.

For simplicity, the surface separation term is modeled as a hard constraint for enforcing the minimum separation between a pair of surfaces as shown in Equation (4).

$$H_a(L_a(S_{i+1}(a)), L_a(S_i(a))) =$$
$$\begin{cases} \infty, & \text{if } L_a(S_{i+1}(a)) - L_a(S_i(a)) < d_{i,i+1} \\ 0, & \text{otherwise} \end{cases}$$ (4)

where $d_{i,i+1}$ is the minimum separation between a pair of adjacent surfaces. The method is also capable of incorporating a convex surface separation penalty while enforcing a minimum separation constraint in the irregularly sampled space using the same framework and is discussed in Section 5.

## 2.2. Graph Construction

For each surface $S_i$, a subgraph $G_i$ is constructed. Herein, the intracolumn edges are added to enforce surface monotonicity and encode the data term for cost volume $D_i$ (for searching $S_i$). Inter-column edges are added between a pair of neighboring columns $a$ and $b$ to enforce the surface smoothness penalty term $V_{ab}(.)$. The graph $G$ for the simultaneous search of all $\lambda$ surfaces consists of the union of those $\lambda$ subgraphs $G_i$'s. Furthermore, inter-surface edges are added between the corresponding columns of subgraphs $G_i$ and $G_{i+1}$ to incorporate the surface separation term for surface distance changes between two surfaces. A pair of columns with respect to the same $(x,y)$ pairs in the domain $\mathbf{x} \times \mathbf{y}$ of subgraphs $G_i$, $G_{i+1}$ for two adjacent surfaces is defined as corresponding columns. The graph $G$ is then solved by computing a maximum flow which minimizes the energy function $E(S)$ (Equation. (2)). The positions of the $\lambda$ target surfaces are obtained by mapping the resultant solution to the physical space using the mapping function $L_a(.)$. The graph is constructed using the cost volumes generated for $\lambda$ surfaces from volume $I(x,y,z)$. Each element in the cost volume $D_i$ to search $S_i$ is represented by a node $n_i(a,z)$ $(z \in \mathbf{z})$ in $G_i$. The following edges are added to incorporate the different energy terms:

## 2.2.1. Intra-column Edges

To ensure the monotonicity of the target surfaces (i.e., the target surface intersects each column exactly one time) and encode the data cost term; intra-column edges are added to each subgraph $G_i$ as described in Ref. Li et al. (2006). Along every column $a$ for surface $S_i$, each node $n_i(a,z)(z > 0)$ has a directed edge with $+\infty$ weight to the node immediately below it and an edge with $D_i(L_a(z-1))$ weight in the opposite direction. Additionally, an edge with $+\infty$ weight is added from the source node $s$ to each node $n_i(a,0)$ and an edge with $D_i(L_a(Z-1))$ weight is added from node $n_i(a,Z-1)$ to the terminal node $t$.

Any $s$-$t$ cut with finite cost contains only one of the finite weight edges $D_i(L_a(.))$ for each column $a$, thus enforcing surface monotonicity. This is because, if any $s$-$t$ cut included more than one finite weight edges, then by construction it must include at least one infinite weight edge thereby making its cost infinite.

### 2.2.2. Inter-column Edges

Inter-column arcs are added between pairs of neighboring columns $a$ and $b$ to each subgraph $G_i$ to encode the surface smoothness term. For the purpose of this paper the incorporation of a convex smoothness term is presented. Denote a function operator $f(r_1, r_2)$ as shown in Equation (5).

$$f(r_1, r_2) = \begin{cases} 0, & \text{if } r_1 < r_2 \\ \psi(r_1 - r_2), & \text{otherwise} \end{cases} \tag{5}$$

where $\psi(.)$ is a convex function.

A general weight setting function $g(.)$ is used for the inter-column edges between two neighboring columns. The following inter-column edges are added:

For all $k_1 \in [0, Z-1]$ and $k_2 \in [1, Z-1]$, a directed edge with weight setting $g(k_1, k_2)$ as shown in Equation (6) is added from node $n_i(a, k_1)$ to node $n_i(b, k_2)$. Additionally, a directed edge is added from node $n_i(a, k_1)$ to terminal node $t$ with weight setting $g(k_1, Z)$.

$$g(k_1, k_2) = f(L_a(k_1), L_b(k_2 - 1)) \\ - f(L_a(k_1 - 1), L_b(k_2 - 1)) - f(L_a(k_1), L_b(k_2)) + f(L_a(k_1 - 1), L_b(k_2)) \tag{6}$$

Where, if $k_1 = 0$, (that is $k_1 - 1 \in/ \mathbf{z}$), then $f(L_a(k_1 - 1), L_b(k_2 - 1)) = f(L_a(k_1 - 1), L_b(k_2)) = 0$ and if $k_2 = Z$, (that is, $k_2 \in/ \mathbf{z}$), then $f(L_a(k_1), L_b(k_2)) = f(L_a(k_1 - 1), L_b(k_2)) = 0$.

**Lemma 1:** For any $k_1$ and $k_2$, the function $g(k_1, k_2)$ is non-negative. (Proof in Appendix A)

In a similar manner, for all $k_1 \in [0, Z-1]$ and $k_2 \in [1, Z-1]$, edges are constructed from nodes $n_i(b, k_1)$ to nodes $n_i(a, k_2)$ with weight setting $g(k_1, k_2)$ as shown in Equation (7). Additionally a directed edge is added from node $n_i(b,k_1)$ to terminal node $t$ with weight setting $g(k_1,Z)$.

$$g(k_1,k_2) = f(L_b(k_1),L_a(k_2-1))$$
$$- f(L_b(k_1-1),L_a(k_2-1)) - f(L_b(k_1),L_a(k_2)) \quad (7) \; + f(L_b(k_1-1),L_a(k_2))$$

It should be noted that the weight setting function $g(k_1,k_2)$ in Equation (7) is similar to Equation (6) with only the column mapping functions $L_a(.)$ and $L_b(.)$ interchanged. Also, in practice we do not add edges with a weight of zero in the graph.

**Lemma 2:** In any finite $s$-$t$ cut $C$, the total weight of the edges between any two adjacent columns $a$ and $b$ (denoted by $C_{a,b}$) equals to the surface smoothness cost of the resulting surface $S_i$ with $S_i(a) = k_1$ and $S_i(b) = k_2$, which is $\psi(L_a(k_1) - L_b(k_2))$, where $\psi(.)$ is a convex function. (Proof in Appendix B)

Example of a graph construction of two neighboring columns $a$ and $b$ for a given surface with enforcement of convex surface smoothness constraint is shown in Fig. 3. Herein, an edge from $n_i(a,k_1)$ to node $n_i(b,k_2)$ is denoted as $E_i(a_{k_1},b_{k_2})$ for the $i$-th surface. For clarity, an edge $E_i(a_{k_1},b_{k_2})$ is denoted as Type I if $k_2 > k_1$, as Type II if $k_2 = k_1$ and as Type III if $k_2 < k_1$. The respective edge weights in the graph are summarized in Table 1. The convex function used in the example is a linear one, taking the form $\psi(k_1 - k_2) = |k_1 - k_2|$.

The following can be verified from the example shown Fig. 3:

- The correct cost of cut $C_1 = |21 - 12| = 9$. It can be verified that the inter-column edges contributing to the cost of cut $C_1$ are Type I edges $E(a_2,b_3)$ and $E(a_1,b_3)$. Summing the edge weights from Table 1, cost of cut $C_1 = 5 + 4 = 9$.

- The correct cost of cut $C_2 = |25 - 37| = 12$. It can be verified that the inter-column edges contributing to the cost of cut $C_2$ are Type I edges $E(b_4,a_5)$, $E(b_3,a_4)$ and Type II edge $E(b_4,a_4)$. Summing the edge weights from Table 1, cost of cut $C_2 = 3 + 3 + 6 = 12$.

- The correct cost of cut $C_3 = |25 - 3| = 22$. It can be verified that the inter-column edges contributing to the cost of cut $C_3$ are Type I edges $E(a_0,b_2)$, $E(a_1,b_2)$, $E(a_1,b_3)$, $E(a_2,b_3)$, Type II edge $E(a_3,b_3)$. Summing the edge weights from Table 1, cost of cut $C_3 = 1 + 8 + 4 + 5 + 4 = 22$.

- The correct cost of cut $C_4 = |25 - 1| = 24$. It can be verified that the inter-column edges contributing to the cost of cut $C_4$ are Type I edges $E(a_0,b_1)$, $E(a_0,b_2)$, $E(a_1,b_2)$, $E(a_1,b_3)$, $E(a_2,b_3)$, Type II edge $E(a_3,b_3)$. Summing the edge weights from Table 1, cost of cut $C_4 =$
  $2 + 1 + 8 + 4 + 5 + 4 = 24$.

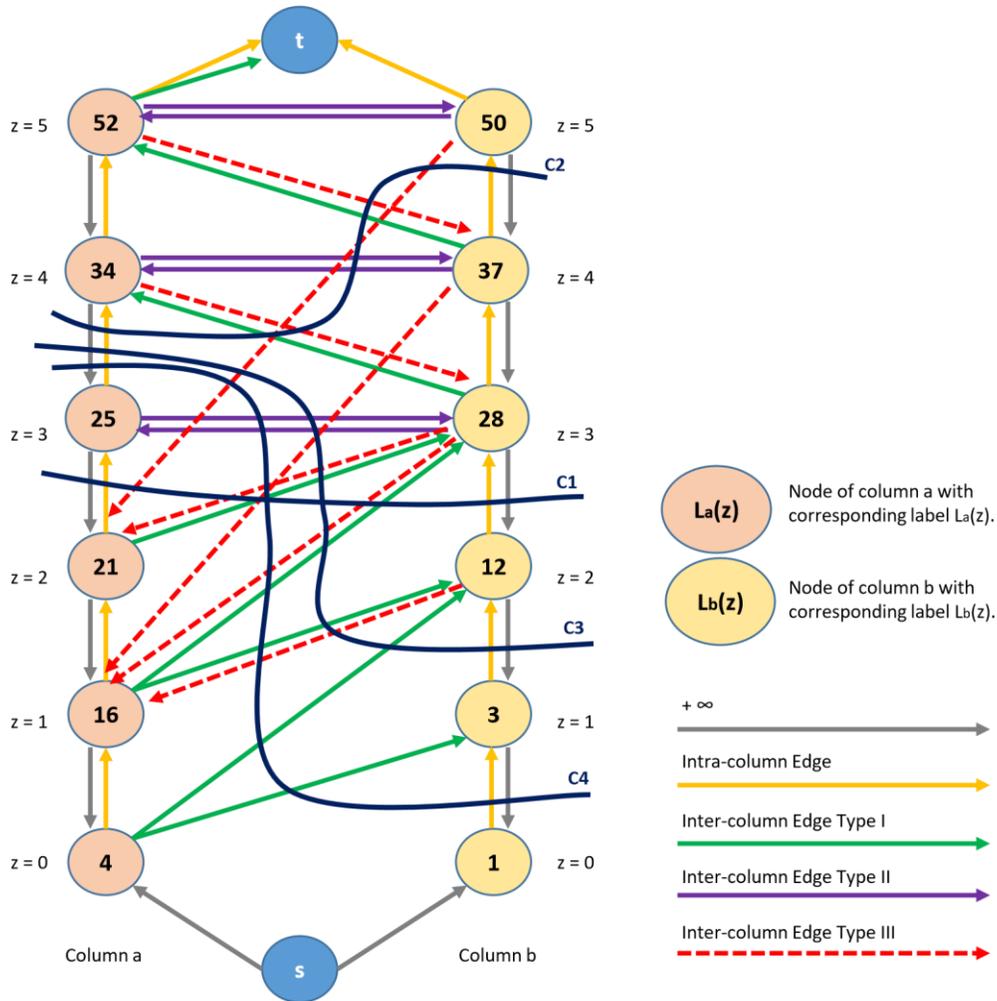

Figure 3: Example graph construction of two neighboring columns $a$ and $b$ to demonstrate enforcement of convex surface smoothness constraints in irregularly sampled space.

Table 1: Summary of inter-column edge weights of the graph construction in Fig. 3, based on a linear function of the form $\psi(k_1 - k_2) = |k_1 - k_2|$

| Edge | Type | Weight | Edge | Type | Weight |
|------|------|--------|------|------|--------|
| $E(a_0,b_1)$ | I | 2 | $E(b_2,a_1)$ | III | 8 |
| $E(a_0,b_2)$ | I | 1 | $E(b_3,a_1)$ | III | 4 |
| $E(a_1,b_2)$ | I | 8 | $E(b_3,a_2)$ | III | 5 |
| $E(a_1,b_3)$ | I | 4 | $E(b_3,a_3)$ | II | 4 |
| $E(a_2,b_3)$ | I | 5 | $E(b_3,a_4)$ | I | 3 |
| $E(a_3,b_3)$ | II | 4 | $E(b_4,a_4)$ | II | 6 |
| $E(a_4,b_3)$ | III | 3 | $E(b_4,a_5)$ | I | 3 |
| $E(a_4,b_4)$ | II | 6 | $E(b_5,a_5)$ | II | 13 |
| $E(a_5,b_4)$ | III | 3 | | | |
| $E(a_5,b_5)$ | II | 13 | | | |
| $E(a_5,b_6)$ | I | 2 | | | |

### 2.2.3. Inter-surface Edges

The surface separation term $H_a(.)$ between two adjacent surfaces is enforced by adding edges in a similar manner as described in Ref. (Abra`moff et al., 2014) from column $a$ in subgraph $G_i$ to corresponding column $a$ in subgraph $G_{i+1}$. Along every column $a$ in $G_i$, each node $n_i(a,z)$ has a directed edge with $+\infty$ weight to the node $n_{i+1}(a,z^0)$, ($z^0 \in$ **z**, $L_a(z^0) - L_a(z) \geq d_{i,i+1}, L_a(z^0 - 1) - L_a(z) < d_{i,i+1}$). Additionally an edge with $+\infty$ weight is added from node $n_i(a,z)$ to the terminal node $t$ if $L_a(Z-1) - L_a(z) < d_{i,i+1}$.

It can be verified, that no finite $s$-$t$ cut is possible when $L_a(z^0) - L_a(z) < d_{i,i+1}$, since by construction an inter-surface edge of $+\infty$ weight will be cut, thus making the cost infinite. An example of a graph construction for two corresponding columns of adjacent pair of surfaces with enforcement of the surface separation constraint is shown in Fig. 4.

Thus the surface separation term $H_a(.)$ is correctly encoded in graph $G$. Note that if $H_a(.)$ is modeled with a convex function, the same graph construction as that for the surface smoothness term can be used to encode it in the graph.

### 2.3. Surface Recovery from Minimum $s$-$t$ cut

The minimum $s$-$t$ cut in the graph then defines optimal $\lambda$ surfaces $S_i$ where $i = 1,2...\lambda$. For a given surface $S_i$, the surface label for each $col(x,y) \in$ **z**, where $x \in$ **x** and $y \in$ **y** is given by the minimum $s$-$t$ cut (Li et al., 2006). The final surface for each recovered by the mapping $\{0, 1, \dots Z-1\}$ positions column $a$ is applying function $L_a :$ → R, where $a \in \mathbf{x} \times \mathbf{y}$, thereby yielding the resultant surface positions for each column $L_a(S_i(a)) \in \tilde{z}$, where $\tilde{z} \in$ R.

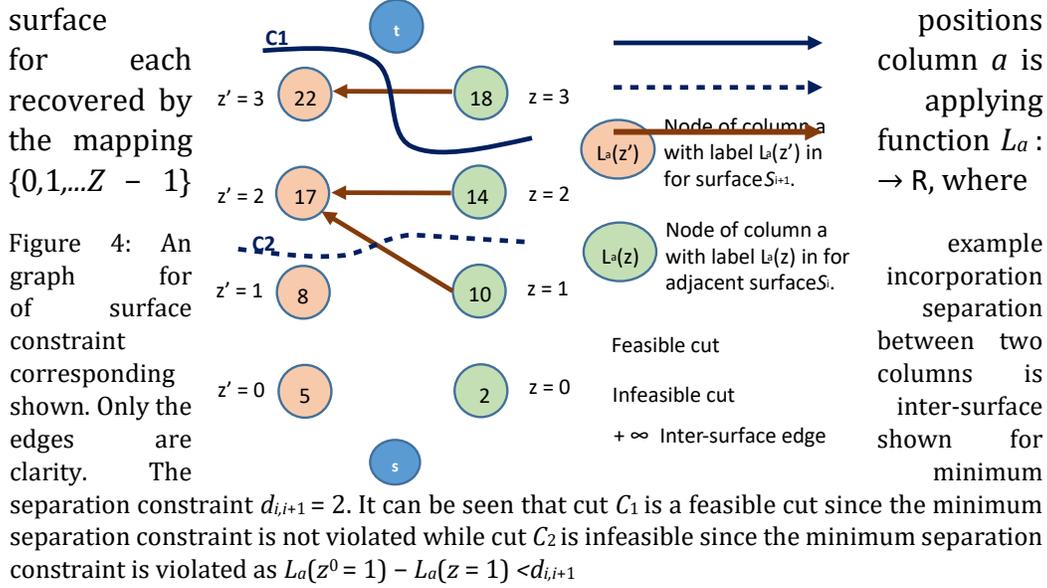

Figure 4: An graph for of surface constraint corresponding shown. Only the edges are clarity. The example incorporation separation between two columns is inter-surface shown for minimum separation constraint $d_{i,i+1} = 2$. It can be seen that cut $C_1$ is a feasible cut since the minimum separation constraint is not violated while cut $C_2$ is infeasible since the minimum separation constraint is violated as $L_a(z^0 = 1) - L_a(z = 1) < d_{i,i+1}$

## 3. Experimental Methods

### 3.1. Intravascular Ultrasound (IVUS) Images

To study the applicability of the proposed method, the segmentation of lumen and media with subvoxel accuracy was performed in Intravascular Ultrasound (IVUS) images as shown in Fig. 5.

Atherosclerosis, a disease of the vessel wall, is the major cause of cardiovascular diseases such as heart attack or stroke (Frosteg°ard, 2005). Early atherosclerosis results in remodeling, thus retaining the lumen despite plaque accumulation (Glagov et al., 1987). Atherosclerosis plaque is located between lumen and media that can be identified in IVUS images. Automated IVUS segmentation of lumen and media is of substantial clinical interest and contributes to clinical diagnosis and assessment of plaque (Balocco et al., 2014).

In this experiment we compare the segmentation accuracy of the lumen and media using the proposed method with the complete set of methods used in the standardized evaluation of IVUS image segmentation (Balocco et al., 2014). The compared methods are namely, P1 - Shape driven segmentation based on linear projections (Unal et al., 2008), P2 - geodesic active contour based segmentation (Caselles et al., 1997), P3 - Expectation maximization based method (Cardinal et al., 2006, 2010), P4 - graph search based method (Downe et al., 2008), P5 - Binary classification of distinguishing between lumen and non-lumen regions based on multi-scale Stacked Sequential learning scheme (Gatta et al., 2011), P6 - Detection of Media border by holistic interpretation of the IVUS image (HoliMAb) (Ciompi et al., 2012), P7 - Lumen segmentation based on a Bayesian approach (Mendizabal-Ruiz et al., 2013), P8 - Sequential detection (Bourantas et al., 2008). Herein, method P4 is based on the optimal surface segmentation method using hard constraints (Li et al., 2006) applied on regularly sampled space. For fair and robust analysis, we also compare the segmentation accuracy of the proposed method in the irregularly sampled space to the optimal surface segmentation method using convex smoothness constraints in the regularly sampled space (OSCS) (Song et al., 2013) and applied deformations to the OSCS segmentation results (DOSCS as described in Section 3.1.2). The proposed method, OSCS and DOSCS method employ the same parameter settings. Additionally, we compare the measures obtained from our method to a deep learning method with a UNET architecture (UNET) which was applied on the same dataset and was reported in Ref. (Balakrishna et al., 2018). Overview of each method's feature, including whether the algorithm was applied to lumen and/or media, whether the segmentation was done in 2-D or 3-D and whether the method was semi-automated or fully automated is shown in Table 2.

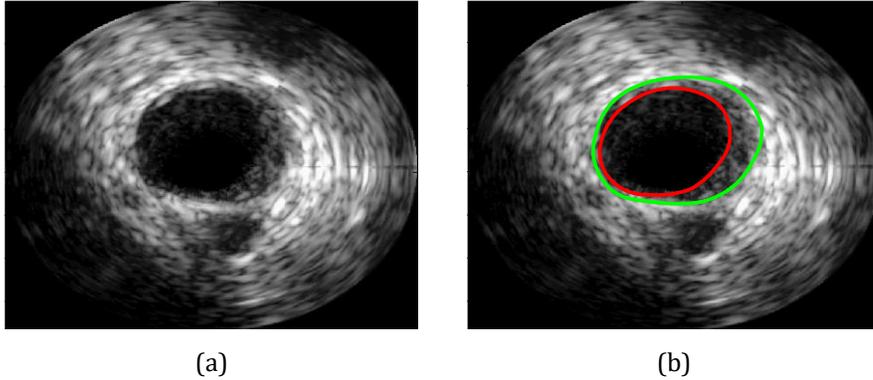

(a)                      (b)

Figure 5: (a) A single frame of an IVUS multiframe dataset (b) Expert manual tracings of the Lumen (red) and Media (green).

Table 2: Overview of the compared method features

| Methods | Category | Automation | 2-D/3-D |
|---|---|---|---|
| P1 (Shape driven) | Lumen and Media | Semi | 2-D |
| P2 (Active contour) | Lumen | Semi | 2-D |
| P3 (Expectation maximization) | Lumen and Media | Semi | 2-D |
| P4 (Graph search) | Lumen and Media | Fully | 3-D |
| P5 (Sequential learning) | Lumen | Fully | 3-D |
| P6 (HoliMAb) | Media | Fully | 2-D |
| P7 (Bayesian) | Lumen | Semi | 2-D |
| P8 (Sequential detection) | Lumen and Media | Fully | 2-D |
| UNET (Deep learning based) | Lumen and Media | Fully | 2-D |
| OSCS | Lumen and Media | Fully | 3-D |
| DOSCS | Lumen and Media | Fully | 3-D |
| Our Method | Lumen and Media | Fully | 3-D |

### 3.1.1.  Data

The data used for this experiment was obtained from the standardized evaluation of IVUS image segmentation (Balocco et al., 2014) database. In this experiment Dataset B as denoted in Ref. Balocco et al. (2014) was used. The data comprises of a set of 435 images with a size of 384 × 384 pixels extracted from in vivo pullbacks of human coronary arteries from 10 patients. The respective expert manual tracings (subvoxel accurate) of lumen and media for the images were also obtained from the reference database. The dataset contains 10 multi-frame datasets, in which 3D context from a full pullback is provided. Each dataset comprises of between 20 and 50 gated frames extracted from the full pullback at the end-diastolic cardiac phase. Further, the obtained data comprised of two groups - training and testing set. Approximately one fourth of the images in the dataset were grouped in the training set and the remaining were grouped as the testing set, to assure fair evaluation of the algorithms with respect to the expert manual tracings. The experiment with the proposed method was conducted in conformance with the directives provided for the IVUS challenge (Balocco et al., 2014).

### 3.1.2.  Workflow

Each slice of the volumes in the dataset is first converted into a polar coordinate image as shown in Fig 6. For each frame, given the center of the

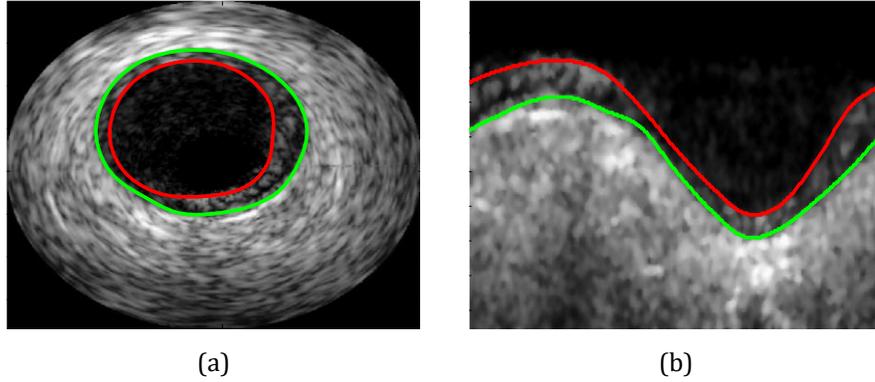

<div align="center">(a)                         (b)</div>

Figure 6: (a) A single frame of an IVUS multiframe dataset (b) Polar transformation of (a). Red - Lumen, Green - Media.

image, for each angular position $\theta = \{0,1,...360\}$ degrees on the short-axis view (Balocco et al., 2014), the corresponding radial columns are generated by considering the gray-level values of the sequence along the radius at the chosen angle and the generated columns are stacked consecutively to generate the polar image volumes. The generated polar image volumes undergo the application of a 7 × 7 × 7 Gaussian filter with a standard deviation of 4 for denoising. Next, the cost image volumes $D_{lumen}$ and $D_{media}$ are generated for the lumen and media respectively. The OSCS method is applied to the cost volumes $D_{lumen}$ and $D_{media}$. Further the GVF as discussed in Section 3.1.3 is computed on the polar image volumes. The deformation field is then applied to cost image volumes and the shifted positions of the voxel centers are recorded. The deformed cost function image volumes $D'_{lumen}$ and $D'_{media}$ are then segmented using the proposed method. The deformation obtained from GVF was applied to the automated segmentations obtained from the OSCS method, resulting in deformed OSCS (DOSCS) segmentations. Finally the resulting segmentations are mapped back to the original coordinate system.

### 3.1.3. Gradient Vector Field

A gradient vector field (GVF) (Xu and Prince, 1998) is a feature preserving diffusion of the gradient in a given image volume. In this study, GVF is used as a deformation field $F(x,y,z)$ obtained directly from the input volume data acting on the center of each voxel $(x,y,z)$ to shift the evenly distributed voxels to the deformed space. The voxel centers are thus displaced towards the regions where salient transitions of image properties are more likely to occur. The shift of the centers of the voxels is given by Equation (8). $(x^0,y^0,z^0)$ = $(x,y,z) + \gamma F(x,y,z)$    (8) where $\gamma$ is a normalization factor. The displacement of each voxel center is confined to the same voxel. Therefore, $F(x,y,z)$ is normalized such that the maximum deformation is equal to half of the voxel size $\delta$. The normalization factor takes the following form as show in Equation (9).

$$\lambda = \frac{\delta}{2 \times max_{(x,y,z)\in(X,Y,Z)}||F(x,y,z)||}$$

(9)

### 3.1.4. Cost Function Design

To detect the lumen and media, a machine learning approach is adopted to generate cost images. For each pixel of the polar image in the training set, a total of 148 features were generated. The following operators are applied in order to generate the features:

- 16 features are generated by applying a set of 16 Gabor filters to the image according to the following kernel shown in Equation (10).

$$G(x,y) = \frac{1}{2\pi\sigma_x\sigma_y}e^{-0.5\times((\frac{x}{\sigma_x})^2+(\frac{y}{\sigma_y})^2)+i2\pi(Ux+Vy)}$$

(10)

The parameters U and V (scaling and orientation) used are U = (0.0442, 0.0884,0.1768,0.3536), V = (0,$\pi$/4,$\pi$/2,3$\pi$/4), $\sigma_x$ = 0.5622U and $\sigma_y$ = 0.4524U .

- 2 features are generated by applying a 3 × 3 Sobel kernel to the image in the $x$ and $y$ directions.

- 6 features are generated by computing the mean value (m), standard deviation (s) and the ratio $\frac{m}{s}$ of pixel intensities in a sliding window of size 1 × 10 pixels in the $x$ and $y$ directions.

- 2 features defined as shadow (Sh) and relative shadow (Sr) related to the cumulative gray level of the image are generated as shown in the following Equations (11),(12).

$$Sh(x,y) = \frac{1}{N_r N_c}\sum_{y_s=y}^{N_r} BI(x,y_s)$$

(11)

$$Sr(x,y) = \frac{1}{N_r N_c} \sum_{y_s=y}^{N_r} y_s BI(x, y_s)$$

(12)

where $BI(x,y)$ is a binary image obtained by thresholding the image with a thresholding value = 14 and ($N_r, N_c$) are the image dimensions.

- 1 feature is generated by computing the local binary pattern (Ojala et al., 2002).

- 121 features are generated by using a 11 × 11 window centered at each pixel in the image, comprising of the intensity values of each pixel in the given window.

Using the expert manual tracings for the training set two separate random forest classifiers (Breiman, 2001) for lumen and media with 10 trees are trained on all the pixels of the images in the training set to learn the probability maps which indicate the likelihood of a pixel belonging to lumen or media respectively. Finally, the trained classifiers are then applied to each pixel of the testing set to obtain the two cost images $D_{lumen}$, $D_{media}$ for lumen and media.

### 3.1.5. Parameter Setting

A linear (convex) function, $\psi(k_1 - k_2) = |k_1 - k_2|$ was used to model the surface smoothness term $V_{ab}(.)$. The surface separation term $H_a(.)$ is modeled as a hard constraint for enforcing the minimum separation between the lumen and media with $d_{lumen,media}$ = 2.

## 4. Results

The quantitative analysis was carried out by comparing the segmentations obtained by the proposed and compared methods with the expert manual tracings (subvoxel accurate). Three evaluation measures were used to quantify the accuracy of the segmentations. The measures used are:

Jaccard Measure (JM) - Quantifies how much the segmented area overlaps with the manual delineated area as shown in Equation (13):

$$JM(R_{auto}, R_{man}) = \frac{|R_{auto} \cap R_{man}|}{|R_{auto} \cup R_{man}|}$$

(13)

where $R_{auto}$ and $R_{man}$ are two vessel regions defined by the manual annotated contour $C_{man}$ and of the automated segmented outline $C_{auto}$ respectively.

Percentage of Area Difference (PAD) - Computes the segmentation area difference as shown in Equation (14) :

$$PAD = \frac{|A_{auto} - A_{man}|}{A_{man}}$$

(14)

where $A_{auto}$ and $A_{man}$ are the vessel areas for the automatic and manual contours respectively.

Hausdroff Distance (HD) - Computes locally the distance between the manual and automated contours as shown in Equation (15).

$$HD(C_{auto}, C_{man}) = max_{p \in C_{auto}}\{max_{q \in C_{man}}[d(p,q)]\}$$ (15) where $p$ and $q$

are points of the curves $C_{auto}$ and $C_{man}$, respectively, and

$d(p,q)$ is the Euclidean distance.

The quantitative results are summarized in Table 3. The results demonstrate that our method performs better than methods P1, P2, P4, P5, P6, P8 and is comparable to methods P3 and P7 with respect to segmentation error measures for lumen and media. Our method segments both the lumen and media simultaneously while method P7 segments the lumen only. Furthermore, our method is fully automated while methods P3 and P7 are semi-automated. Finally, methods P3 and P7 perform slice by slice segmentation in 2-D while our method performs the segmentation in 3-D and not slice by slice.

For the UNET method (Balakrishna et al., 2018), the authors published the performance of their method with respect to Jaccard Metric. It can be seen from the results that based on the Jaccard metric, the proposed method outperforms the UNET method.

The quantitative results also show that the proposed method yields more accurate segmentations than the OSCS and DOSCS methods for both the Lumen and the Media surfaces. The JM obtained from the segmentation results by our proposed method were significantly higher (p<0.01) than the JM computed with the segmentation results from the OSCS and the DOSCS methods. The PAD and HD metrics computed with the proposed method were significantly lower (p<0.01) than the PAD and HD metrics computed with the segmentation results from the OSCS and the DOSCS methods. We did not have access to the actual segmentation results from the P1-P8 methods to perform a paired t-test for significance determination and to qualitatively compare the segmentation results.

Table 3: Evaluation measures of each method with respect to expert manual tracings. Error measures expressed as mean and (standard deviation). An empty table cell indicates that the method was not applied to Lumen or Media. OM-Our Method

| Methods | Lumen | | | Media | | |
|---|---|---|---|---|---|---|
| | JM | PAD | HD | JM | PAD | HD |
| P1 | 0.81 (0.12) | 0.14 (0.13) | 0.47 (0.39) | 0.76 (0.13) | 0.21 (0.16) | 0.64 (0.48) |
| P2 | 0.83 (0.08) | 0.14 (0.12) | 0.51 (0.25) | | | |
| P3 | **0.88 (0.05)** | **0.06 (0.05)** | **0.34 (0.14)** | **0.91 (0.04)** | **0.05 (0.04)** | **0.31 (0.12)** |
| P4 | 0.77 (0.09) | 0.15 (0.12) | 0.47 (0.22) | 0.74 (0.17) | 0.23 (0.19) | 0.76 (0.48) |
| P5 | 0.79 (0.08) | 0.16 (0.09) | 0.46 (0.30) | | | |
| P6 | | | | 0.84 (0.10) | 0.12 (0.12) | 0.57 (0.39) |
| P7 | 0.84 (0.08) | 0.11 (0.12) | 0.38 (0.26) | | | |
| P8 | 0.81 (0.09) | 0.11 (0.11) | 0.42 (0.22) | 0.79 (0.11) | 0.19 (0.19) | 0.60 (0.28) |
| UNET | 0.80 () | | | 0.81 () | | |
| OSCS | 0.80 (0.09) | 0.13 (0.07) | 0.43 (0.19) | 0.81 (0.08) | 0.11 (0.14) | 0.51 (0.19) |
| DOSCS | 0.82 (0.08) | 0.12 (0.07) | 0.41 (0.17) | 0.84 (0.06) | 0.10 (0.14) | 0.48 (0.16) |
| OM | **0.86 (0.04)** | **0.09 (0.03)** | **0.37 (0.14)** | **0.90 (0.03)** | **0.07 (0.03)** | **0.43 (0.12)** |

The average computation time was 105.48 seconds for the OSCS method, 135.27 seconds for the DOSCS method and 187.35 seconds for the proposed method. The increase in average computation time for the DOSCS method as compared to the OSCS method is because the DOSCS method requires additional steps of computing the deformation and applying the computed deformation to the OSCS solution. The increased computation time of the proposed method as compared to the OSCS and DOSCS method is attributed to the increase in the complexity of the graph which results in higher computation time. For the general convex smoothness function $\psi()$, the constructed graphs for the OSCS and the proposed method have the same number of nodes and edges, that is, each node in a given column has an edge to every node in each of its neighboring columns. In our IVUS experiments, we used a special smoothness function $\psi(d) = |d|$. Thus, in the OSCS graph construction, the weight of many of those edges became 0, which were not necessary to be kept in the graph; while in the graph for the proposed method, there were more non-zero weighted edges. Hence, we observed the increase of computation time for the proposed method over OSCS.

Qualitative results are shown in Fig 7 and 8. Fig 7 demonstrates that our method produced very good segmentation of the lumen and the media. It can also be seen from the illustration that the segmentations from our method are consistent for varying shapes of the lumen and media. Fig 8 shows the comparison of OSCS, DOSCS and the proposed method for lumen and media segmentation. It can be seen from the illustration that the DOSCS method improves upon the OSCS method by applying the deformation to the OSCS segmentation results, while the proposed method achieves more accuracy than DOSCS for both lumen and media. Constructing the graph with the shifted voxel centers provides a more accurate encoding of the lumen and media surface positions due to the application of the GVF by adaptively changing the regional node density so that it is higher in regions where the target surface is expected to pass through. Employing a subvoxel accuracy approach allows the segmentation to obtain a higher precision with respect to the OSCS and DOSCS method segmentations.

## 5. Discussion

A novel approach for segmentation of multiple surfaces with convex priors in irregularly sampled space (non-equidistant spacing between orthogonal adjoining nodes) was proposed. Our method advances the graph based segmentation framework in several important ways. First, the proposed energy function incorporates a convex surface smoothness penalty in irregularly sampled space through a convex function. Second, the approach allows simultaneous segmentation of multiple surfaces in the irregularly sampled space with the enforcement of a minimum separation constraint. Third, our method guarantees global optimality. Lastly, the proposed method demonstrates utility in achieving subvoxel segmentation accuracy while employing a convex penalty to model surface smoothness. To the best of our knowledge, this is the first method that fulfills these four aims at the same time. The hallmark of the proposed method is the ability to perform the segmentation task in an irregularly sampled space which generalizes the optimal surface segmentation framework. The proposed method was employed in rapid fat water segmentation in MRI images and demonstrated increased efficiency and accuracy (Cui et al., 2018).

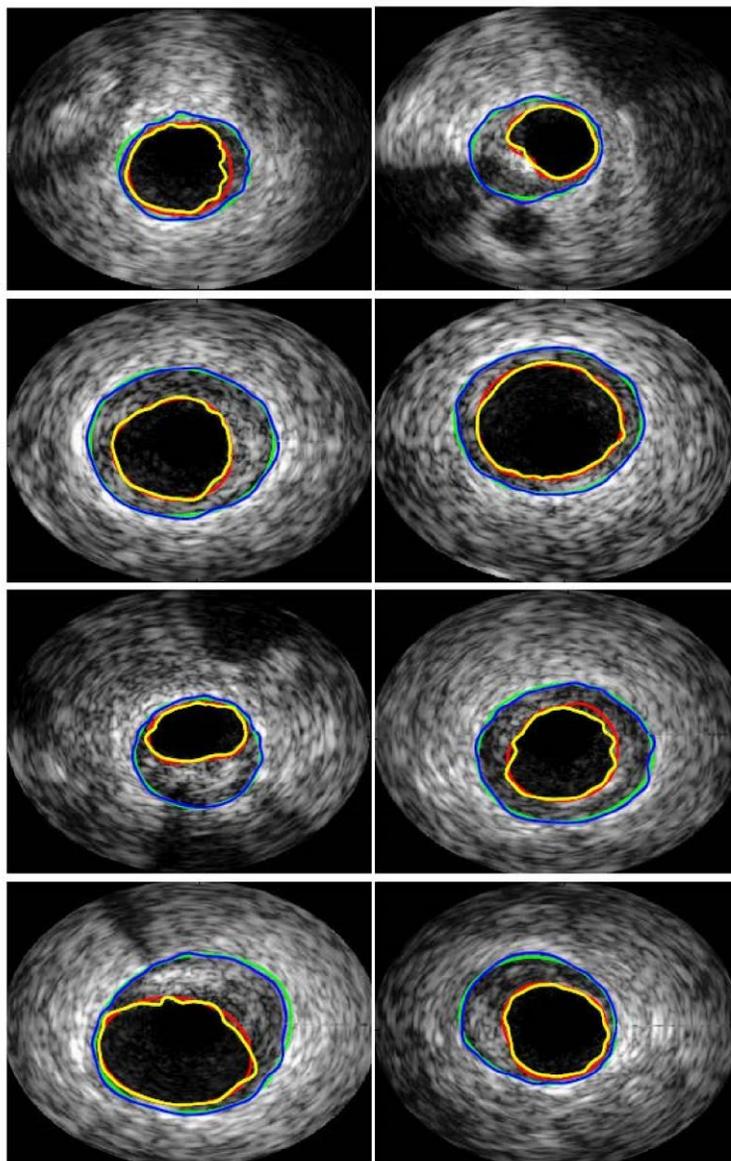

Figure 7: Qualitative illustrations of lumen and media segmentation using our method.

Each image is a single frame of an IVUS multiframe dataset. Red - Lumen expert tracing, Green - Media expert tracing, Yellow - Lumen segmentation (our method), Blue - Media segmentation (our method).

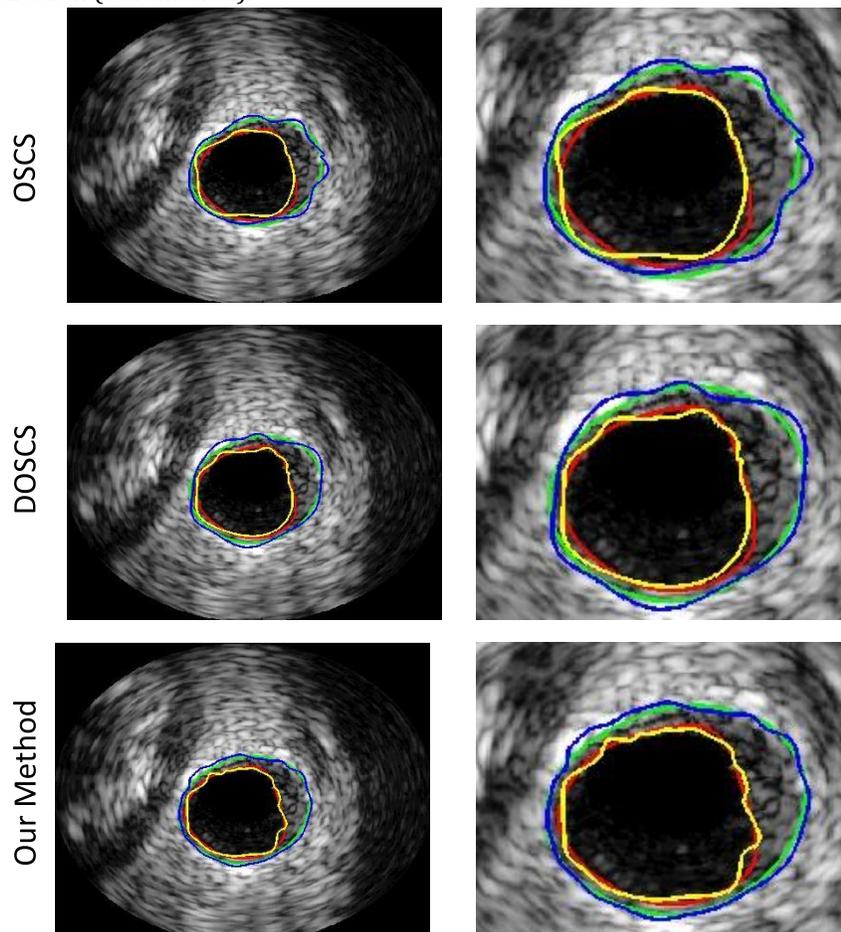

Figure 8: Qualitative illustrations of lumen and media segmentation using OSCS, DOSCS and our method. The first column shows the same single frame of an IVUS multiframe dataset. The second column shows a magnified version of the lumen and media segmentation for each compared method. Red - Lumen expert tracing, Green - Media expert tracing, Yellow - Automated lumen segmentation, Blue - Automated media segmentation.

The proposed method is also capable of incorporating convex surface separation penalty while enforcing a minimum separation in the irregularly sampled space. The incorporation of such a penalty would involve modifying the surface separation term in the proposed energy function to impose a convex function based penalty when the minimum separation constraint is not violated. The graph construction to enforce such a penalty can be done using the same framework of the proposed method for enforcing the surface smoothness constraint.

The method can be used in conjunction with the method proposed by Abra´moff et. al (Abr`amoff et al., 2014) to incorporate prior information using trained hard and soft constraints (Dufour et al., 2013) to achieve subvoxel accuracy. Furthermore, the method can also be incorporated in the image segmentation framework using truncated convex priors (Shah et al., 2015) to achieve subvoxel accuracy by constructing the convex part of the graph in the irregularly sampled space, thus providing a potential use for generic modeling of variety of surface constraints to achieve subvoxel accuracy.

The improved segmentation quality of the proposed method is evident from the illustration in Fig. 8, and shows that segmentation performed in the irregularly sampled space based on the displacement of the voxel centers to correctly encode the partial volume information is more accurate compared to the segmentation performed without any use of partial volume information. The results on IVUS images demonstrates that the methods achieves high accuracy with respect to subvoxel accurate expert tracings as compared to the methods reported in the IVUS challenge (Balocco et al., 2014) while being fully automated and performing segmentation in 3-D. The approach is not limited to these two modalities for which the experiments were conducted.

The proposed method is designed for segmentation problems wherein column structures contain non-equidistant spacing between consecutive elements. Specifically, for subvoxel image segmentation tasks, the voxels centers are deformed. The deformation results in decreased spacing between consecutive voxel centers along a column in certain areas and likewise, increased spacing between voxel centers in certain regions. This creates subvoxel resolution in areas with decreased spacing while super-voxel resolution in areas with increased spacing between the voxel centers. The effect of the supervoxel resolution in those areas is alleviated due to subvoxel resolution in areas containing voxels with high likelihood for presence of the surface boundary.

Recently, deep learning methods have also been extensively used in various medical image analysis and segmentation applications (Litjens et al., 2017). However, deep learning algorithms are inherently limited to amount of training data and corresponding availability of expert annotated truth. While the proposed method is capable of performing subvoxel-accurate segmentations, majority of the deep learning methods are applied at a voxel level segmentation/classification tasks. The result from the UNET method demonstrated the superior performance of the proposed method over traditional deep learning methods. However, it should be noted that the UNET method was applied in 2-D while UNETs can also be applied in 3-D, which may result in improvement of results. Furthermore, many more sophisticated 2-D/3-D deep learning methods such as conditional GANs have recently been developed and have shown to achieve high accuracy in segmentation tasks. Application of such state-of-the-art deep learning methods may also result in improvement of segmentation performance.

## 6. Conclusion

We presented a general framework for simultaneous segmentation of multiple surfaces in the irregularly sampled space with convex priors to achieve subvoxel and super resolution segmentation accuracy. An edge-weighted graph representation is presented and a globally optimal solution with respect to the employed objective function is achieved by solving a maximum flow problem. The surface smoothness and surface separation constraints provide a flexible means for modeling various inherent properties and interrelations of the desired surfaces in an irregularly sampled grid space. The method is readily extensible to higher dimensions.

## Appendix A

**Lemma 1:** For any $k_1$ and $k_2$, the function $g(k_1,k_2)$ is non-negative.

**Proof:** Let us consider the function $g(k_1,k_2)$ for edges from column $a$ to neighboring column $b$ as shown in Equation (6). We need to prove that $g(k_1,k_2) \geq 0$

$$g(k_1,k_2) = f(L_a(k_1),L_b(k_2-1))$$
$$- f(L_a(k_1-1),L_b(k_2-1)) - f(L_a(k_1),L_b(k_2))$$
$$+ f(L_a(k_1-1),L_b(k_2))$$

The reader should recall because of the strictly increasing order of sampling, $L_a(k_1) > L_a(k_1-1)$ and $L_b(k_2) > L_b(k_2-1)$. $\psi(\cdot)$ is a convex function with $\psi(0) = 0$. The proof is presented in a case-by-case basis.

**Case 1:** $L_a(k_1) < L_b(k_2 - 1)$

Thus, $L_a(k_1 - 1) < L_b(k_2 - 1)$. As $L_b(k_2) > L_b(k_2 - 1)$, we have $L_a(k_1) < L_b(k_2)$ and $L_a(k_1 - 1) < L_b(k_2)$. Since $f(r_1, r_2) = 0$ if $r_1 < r_2$. It is straightforward to verify that $g(k_1, k_2) = 0$ in Equation (6).

**Case 2:** $L_a(k_1) \geq L_b(k_2 - 1)$ and $L_a(k_1) < L_b(k_2)$

In this case, as $L_a(k_1) > L_a(k_1 - 1)$, we have $L_a(k_1 - 1) < L_b(k_2)$. Thus, $g(k_1, k_2)$ takes the following form in Equation (6).

$$g(k_1, k_2) = f(L_a(k_1), L_b(k_2 - 1)) - f(L_a(k_1 - 1), L_b(k_2 - 1))$$

If $L_a(k_1 - 1) < L_b(k_2 - 1)$, then $g(k_1, k_2) = f(L_a(k_1), L_b(k_2 - 1)) = \psi(L_a(k_1) - L_b(k_2 - 1))$. Thus, $g(k_1, k_2) \geq 0$ as $\psi(L_a(k_1) - L_b(k_2 - 1)) \geq 0$ with $L_a(k_1) \geq L_b(k_2 - 1)$.

If $L_a(k_1 - 1) < L_b(k_2 - 1)$, then $g(k_1, k_2) = \psi(L_a(k_1) - L_b(k_2 - 1)) - \psi(L_a(k_1 - 1) - L_b(k_2 - 1))$. We know that $L_a(k_1) - L_b(k_2 - 1) > L_a(k_1 - 1) - L_b(k_2 - 1) > 0$. Thus, $g(k_1, k_2) > 0$ as $\psi(0) = 0$. Therefore, in this case $g(k_1, k_2) > 0$.

**Case 3:** $L_a(k_1) \geq L_b(k_2)$

In this case, $L_a(k_1) > L_b(k_2 - 1)$ as $L_b(k_2) > L_b(k_2 - 1)$. We distinguish three subcases: 1) $L_a(k_1 - 1) < L_b(k_2 - 1)$, 2) $L_a(k_1 - 1) < L_b(k_2)$ and $L_a(k_1 - 1) \geq L_b(k_2 - 1)$, and 3) $L_a(k_1 - 1) \geq L_b(k_2)$.

**Subcase 1):** If $L_a(k_1 - 1) < L_b(k_2 - 1)$, then $g(k_1, k_2) = f(L_a(k_1), L_b(k_2 - 1)) - f(L_a(k_1), L_b(k_2))$

$$= \psi(L_a(k_1) - L_b(k_2 - 1)) - \psi(L_a(k_1) - L_b(k_2))$$

Since $L_b(k_2 - 1) < L_b(k_2)$, we have $L_a(k_1) - L_b(k_2 - 1) > L_a(k_1) - L_b(k_2)$. Thus, $g(k_1, k_2) > 0$ as $\psi(0) = 0$.

**Subcase 2):** If $L_a(k_1 - 1) < L_b(k_2)$ and $L_a(k_1 - 1) \geq L_b(k_2 - 1)$, then $g(k_1, k_2)$ takes the form shown in Equation (6) as $L_a(k_1) \geq L_b(k_2) > L_a(k_1 - 1) \geq L_b(k_2 - 1)$.

$$g(k_1, k_2) = f(L_a(k_1), L_b(k_2 - 1))$$

$$- f(L_a(k_1 - 1), L_b(k_2 - 1)) - f(L_a(k_1), L_b(k_2))$$

$$= \psi(L_a(k_1) - L_b(k_2 - 1))$$

$$- \psi(L_a(k_1 - 1) - L_b(k_2 - 1)) - \psi(L_a(k_1) - L_b(k_2))$$

Let $L_a(k_1) - L_b(k_2) = \delta_1$, $L_b(k_2) - L_a(k_1 - 1) = \delta_2$ and $L_a(k_1 - 1) - L_b(k_2 - 1) = \delta_3$, where $\delta_1 \geq 0$, $\delta_2 > 0$ and $\delta_3 \geq 0$. Rewriting Equation (6) and substituting these values, we get the following expression expression, $g(k_1, k_2) = \psi(L_a(k_1) - L_b(k_2 - 1))$

$$- \psi(L_a(k_1 - 1) - L_b(k_2 - 1)) - \psi(L_a(k_1) - L_b(k_2))$$

$$= \psi(\delta_1 + \delta_2 + \delta_3) - \psi(\delta_3) - \psi(\delta_1)$$

It can be verified that $g(k_1, k_2) > 0$ as $\psi(\cdot)$ is convex.

**Subcase 3):** If $L_a(k_1 - 1) \geq L_b(k_2)$, then $L_a(k_1) - L_b(k_2 - 1) > 0$, $L_a(k_1 - 1) - L_b(k_2) \geq 0$, $L_a(k_1 - 1) - L_b(k_2 - 1) > 0$, and $L_a(k_1) - L_b(k_2) > 0$. Hence,

$$g(k_1, k_2) = \psi(L_a(k_1) - L_b(k_2 - 1))$$
$$- \psi(L_a(k_1 - 1) - L_b(k_2 - 1)) - \psi(L_a(k_1) - L_b(k_2)) +$$
$$\psi(L_a(k_1 - 1) - L_b(k_2)).$$

In this subcase, let $L_a(k_1) - L_a(k_1 - 1) = \delta_1$, $L_a(k_1 - 1) - L_b(k_2) = \delta_2$ and $L_b(k_2) - L_b(k_2 - 1) = \delta_3$, where $\delta_1 > 0$, $\delta_2 \geq 0$ and $\delta_3 > 0$. Substituting this in the expression for $g(k_1, k_2)$, we get

$$g(k_1, k_2) = \psi(\delta_1 + \delta_2 + \delta_3) - \psi(\delta_2 + \delta_3) - \psi(\delta_1 + \delta_2) + \psi(\delta_2).$$

Let us first consider the case, $\delta_2 = 0$, we get the following expression,

$$g(k_1, k_2) = \psi(\delta_1 + \delta_3) - \psi(\delta_3) - \psi(\delta_1)$$

It can be verified that $g(k_1, k_2) > 0$ as $\psi(\cdot)$ is convex.

Next, consider the case when $\delta_2 > 0$. It can be observed that $\delta_1 + \delta_2 + \delta_3 > \delta_1 + \delta_2 > \delta_2$. Therefore, $\delta_1 + \delta_2$ can be expressed as, $\delta_1 + \delta_2 = \lambda_1 \delta_2 + (1 - \lambda_1)(\delta_1 + \delta_2 + \delta_3)$

Solving for $\lambda_1$, we get $\lambda_1 = \frac{\delta_3}{\delta_1 + \delta_3}$.

Similarly, it can be observed that $\delta_1 + \delta_2 + \delta_3 > \delta_2 + \delta_3 > \delta_2$ and $\delta_2 + \delta_3$ can be expressed as, $\delta_2 + \delta_3 = \lambda_2 \delta_2 + (1 - \lambda_2)(\delta_1 + \delta_2 + \delta_3)$ , where $\lambda_2 = \frac{\delta_1}{\delta_1 + \delta_3}$.

From the definition of a convex function, and adding the above two expressions, we get the following,

$$\psi(\delta_1 + \delta_2) + \psi(\delta_2 + \delta_3) \leq (\lambda_1 + \lambda_2)\psi(\delta_2) + (2 - \lambda_1 - \lambda_2)\psi(\delta_1 + \delta_2 + \delta_3).$$

Substituting the value of $\lambda_1$ and $\lambda_2$, we get $\psi(\delta_1 + \delta_2) + \psi(\delta_2 + \delta_3) \leq \psi(\delta_2) + \psi(\delta_1 + \delta_2 + \delta_3)$. Therefore it can be verified that $g(k_1, k_2) \geq 0$.

Thus, through these exhaustive cases, it is shown that for any $k_1$ and $k_2$, the function $g(k_1, k_2) \geq 0$ or in other words is non-negative.

## Appendix B

**Lemma 2:** In any finite $s$-$t$ cut $C$, the total weight of the edges between any two adjacent columns $a$ and $b$ (denoted by $C_{a,b}$) equals to the surface smoothness cost of the resulting surface $S_i$ with $S_i(a) = k_1$ and $S_i(b) = k_2$, which is $\psi(L_a(k_1) - L_b(k_2))$, where $\psi(.)$ is a convex function.

**Proof:** Denote an edge from $n_i(a, k_1)$ to node $n_i(b, k_2)$ as $E_i(a_{k_1}, b_{k_2})$ for the $i$-th surface. Assume $k_1 \geq k_2$. Proof for the case when $k_2 \geq k_1$ can be done in a similar manner by interchanging the notations for column $a$ and column $b$.

To show: cost of cut $C_{a,b} = \psi(L_a(k_1) - L_b(k_2))$.

We start by observing such a $s$-$t$ cut $C_{a,b}$ will consist of only the following inter-column edges:

$$\{E_i(a_m, b_n)\, ,\, 0 \leq m \leq k_1,\, k_2 + 1 \leq n \leq Z\}$$

Note, here we use the index $Z$ to denote the terminal node $t$ as described in Section 2.2.2.

Summing up the weights of the above edges using Equation 6, we obtain the following expression:

$$
\begin{aligned}
C_{a,b} = &\, g(k_1, Z) + g(k_1, Z - 1) + g(k_1, Z - 2) \\
&+ \ldots + g(k_1, k_2 + 1) \\
&+ g(k_1 - 1, Z) + g(k_1 - 1, Z - 1) + g(k_1 - 1, Z - 2) \\
&+ \ldots + g(k_1 - 1, k_2 + 1)
\end{aligned}
$$

$$\cdot$$
$$\cdot$$
$$\cdot$$
$$+ g(0,Z) + g(0,Z - 1) + g(0,Z - 2)$$
$$+ \ldots + g(0,k_2 + 1)$$

Let us first evaluate part of Equation (6) for $k$, where $0 \leq k \leq k_1$ as shown below: $g(k,Z) + g(k,Z - 1) + g(k,Z - 2) + \ldots + g(k,k_2 + 1)$

$= f(L_a(k),L_b(Z - 1)) - f(L_a(k - 1),L_b(Z - 1))$
$-f(L_a(k),L_b(Z)) + f(L_a(k - 1),L_b(Z))$
$+f(L_a(k),L_b(Z - 2)) - f(L_a(k - 1),L_b(Z - 2))$
$-f(L_a(k),L_b(Z - 1)) + f(L_a(k - 1),L_b(Z - 1))$
$+f(L_a(k),L_b(Z - 3)) - f(L_a(k - 1),L_b(Z - 3))$
$-f(L_a(k),L_b(Z - 2)) + f(L_a(k - 1),L_b(Z - 2))$
$\cdot$
$\cdot$
$\cdot$
$+f(L_a(k),L_b(k_2)) - f(L_a(k - 1),L_b(k_2))$
$-f(L_a(k),L_b(k_2 + 1)) + f(L_a(k - 1),L_b(k_2 + 1))$

$= f(L_a(k),L_b(k_2)) - f(L_a(k - 1),L_b(k_2))$
$-f(L_a(k),L_b(Z)) + f(L_a(k - 1),L_b(Z))$

As described in Section 2.2.2, $f(L_a(k),L_b(Z)) = 0$, $f(L_a(k - 1),L_b(Z)) = 0$ ($\because Z \notin \mathbf{z}$)
$= f(L_a(k),L_b(k_2)) - f(L_a(k-1),L_b(k_2))$   ($B1$) By simplifying Equation (6) using Equation (B1), it follows that:

$$C_{a,b} = f(L_a(k_1),L_b(k_2)) - f(L_a(k_1 - 1),L_b(k_2)) + f(L_a(k_1 - 1),L_b(k_2)) - f(L_a(k_1 - 2),L_b(k_2))$$
$$\cdot$$
$$\cdot$$
$$\cdot$$
$$+ f(L_a(1),L_b(k_2)) - f(L_a(0),L_b(k_2))$$

$$+ f(L_a(0), L_b(k_2)) - f(L_a(-1), L_b(k_2))$$

$$= f(L_a(k_1), L_b(k_2)) - f(L_a(-1), L_b(k_2)) \text{ As}$$

described in Section 2.2.2,

$$f(L_a(-1), L_b(k_2)) = 0, \ (\because -1 \in / \mathbf{z})$$

$$= \psi(L_a(k_1) - L_b(k_2)), \text{ Using Equation(5)}$$

Therefore, for this case it is shown that cost of cut $C_{a,b} = \psi(L_a(k_1) - L_b(k_2))$.

In a similar manner when $k_2 \geq k_1$, the *s-t* cut $C_{b,a}$ will consist of the following inter-column edges:

$$\{E_i(b_m, a_n), \ 0 \leq m \leq k_2, \ k_1 + 1 \leq n \leq Z\}$$

Summing up the weights of the above edges using Equation 7, we obtain the following expression:

$$C_{b,a} = g(k_2, Z) + g(k_2, Z - 1) + g(k_2, Z - 2)$$

$$+ \dots + g(k_2, k_1 + 1) \ g(k_2 - 1, Z) + g(k_2 - 1, Z - 1) + g(k_2$$

$$- 1, Z - 2)$$

$$+ \dots + g(k_2 - 1, k_1 + 1)$$

.

.

.

$$g(0, Z) + g(0, Z - 1) + g(0, Z - 2)$$

$$+ \dots + g(0, k_1 + 1)$$

Similar to the previous case, let us first evaluate part of Equation (7) for $k$, where $0 \leq k \leq k_2$ as shown below: $g(k, Z) + g(k, Z - 1) + g(k, Z - 2) + \dots + g(k, k_1 + 1)$

$$= f(L_b(k), L_a(Z - 1)) - f(L_b(k - 1), L_a(Z - 1)) - f(L_b(k), L_a(Z)) + f(L_b(k - 1), L_a(Z))$$

$+f(L_b(k),L_a(Z-2)) - f(L_b(k-1),L_a(Z-2))$

$-f(L_b(k),L_a(Z-1)) + f(L_b(k-1),L_a(Z-1))$

$+f(L_b(k),L_a(Z-3)) - f(L_b(k-1),L_a(Z-3))$

$-f(L_b(k),L_a(Z-2)) + f(L_b(k-1),L_a(Z-2))$

.

.

.

$+f(L_b(k),L_a(k_1)) - f(L_b(k-1),L_a(k_1))$

$-f(L_b(k),L_a(k_1+1)) + f(L_b(k-1),L_a(k_1+1))$

$= f(L_b(k),L_a(k_1)) - f(L_b(k-1),L_a(k_1))$

$-f(L_b(k),L_a(Z)) + f(L_b(k-1),L_a(Z))$

As described in Section 2.2.2, $f(L_b(k),L_a(Z)) = 0$, $f(L_b(k-1),L_a(Z)) = 0$ ($\because Z \in\!\!\!/ \mathbf{z}$)

$= f(L_b(k),L_a(k_1)) - f(L_b(k-1),L_a(k_1))$   $(B2)$ By simplifying Equation (6) using Equation (B2), it follows that:

$$C_{b,a} = f(L_b(k_2), L_a(k_1)) - f(L_b(k_2-1), L_a(k_1))$$
$$+ f(L_b(k_2-1), L_a(k_1)) - f(L_b(k_2-2), L_a(k_1))$$
$$.$$
$$.$$
$$.$$
$$+ f(L_b(1), L_a(k_1)) - f(L_b(0), L_a(k_1))$$
$$+ f(L_b(0), L_a(k_1)) - f(L_b(-1), L_a(k_1))$$

$= f(L_b(k_2),L_a(k_1)) - f(L_b(-1),L_a(k_1))$ As described in Section 2.2.2, $f(L_b(-1),L_a(k_1)) = 0$, ($\because -1 \in\!\!\!/ \mathbf{z}$)

$= \psi(L_b(k_2) - L_a(k_1))$, Using Equation(5)

Therefore, for this case it is shown that cost of cut $C_{b,a} = \psi(L_b(k_2) - L_a(k_1))$.

This completes the proof.